\documentclass[10pt,twocolumn,letterpaper]{article}

\usepackage{cvpr}              

\usepackage{graphicx}
\usepackage{amsmath}
\usepackage{amssymb}
\usepackage{booktabs}
\usepackage{algorithmic}

\usepackage[ruled]{algorithm2e} 

\usepackage[pagebackref,breaklinks,colorlinks]{hyperref}

\usepackage[capitalize]{cleveref}
\crefname{section}{Sec.}{Secs.}
\Crefname{section}{Section}{Sections}
\Crefname{table}{Table}{Tables}
\crefname{table}{Tab.}{Tabs.}

\def\cvprPaperID{*****} 
\def\confName{CVPR}
\def\confYear{2022}

\begin{document}

\title{Sketch-Guided Scene Image Generation}

\author{Tianyu Zhang\\
JAIST\\
Ishikawa, Japan\\
{\tt\small s2320434@jaist.ac.jp}
\and
Xiaoxuan Xie\\
JAIST\\
Ishikawa, Japan\\
{\tt\small s2310069@jaist.ac.jp}
\and
Xusheng Du\\
JAIST\\
Ishikawa, Japan\\
{\tt\small s2320034@jaist.ac.jp}
\and
Haoran Xie\\
JAIST\\
Ishikawa, Japan\\
{\tt\small xie@jaist.ac.jp}
}

\maketitle

\begin{abstract}
  Text-to-image models are showcasing the impressive ability to create high-quality and diverse generative images. Nevertheless, the transition from freehand sketches to complex scene images remains challenging using diffusion models. In this study, we propose a novel sketch-guided scene image generation framework, decomposing the task of scene image scene generation from sketch inputs into object-level cross-domain generation and scene-level image construction. We employ pre-trained diffusion models to convert each single object drawing into an image of the object, inferring additional details while maintaining the sparse sketch structure. In order to maintain the conceptual fidelity of the foreground during scene generation, we invert the visual features of object images into identity embeddings for scene generation. In scene-level image construction, we generate the latent representation of the scene image using the separated background prompts, and then blend the generated foreground objects according to the layout of the sketch input. To ensure the foreground objects' details remain unchanged while naturally composing the scene image, we infer the scene image on the blended latent representation using a global prompt that includes the trained identity tokens. Through qualitative and quantitative experiments, we demonstrate the ability of the proposed approach to generate scene images from hand-drawn sketches surpasses the state-of-the-art approaches.
\end{abstract}

\section{Introduction}

\begin{figure*}
  \includegraphics[width=\textwidth]{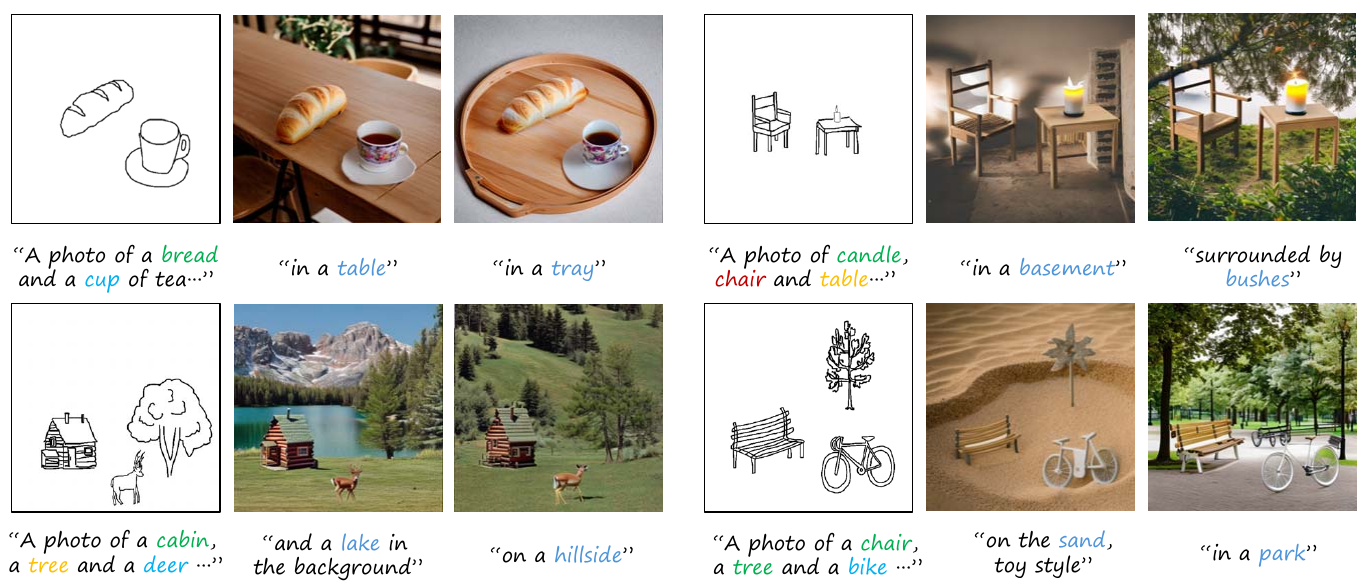}
  \caption{We present a cross-domain generation method from scene sketches to images. The results demonstrate that our method can generate complete semantic foreground and background, maintaining consistency with the input sketches and semantics.}
  \label{fig:teaser}
\end{figure*}

Text-to-image diffusion models~\cite{ho2020denoising,rombach2022high} have significantly enhanced the quality of image generation, demonstrating more robust semantic understanding and content creation capabilities in generative models. However, the complex scene image generation with multiple objects remains a challenging task for diffusion models. When dealing with combinations of multiple objects, diffusion models often encounter catastrophic identity loss and semantic blending issues~\cite{chefer2023attend}. This issue is particularly pronounced when attempting to describe complex semantics using text prompt, such as shape contours and spatial relationships. To solve this issue, the current solutions may involve leveraging external guidance such as semantic masks~\cite{bar2023multidiffusion,zhang2023adding}, layouts~\cite{li2023gligen,chen2024training,phung2023grounded}, and keypoints~\cite{gu2024mix,ju2023human}. These external conditions can guide image generation to achieve the desired outcomes with improved coherence and accuracy. Compared with these conditions, freehand sketches often offer an intuitive and detailed expression of the user's intent~\cite{gao2020sketchycoco}. Particularly when describing an object or scene, sketches can articulate detailed semantic information such as shapes, locations, and relationships, contributing to the construction of comprehensive semantic details.

\begin{figure}
    \centering
    \includegraphics[width=0.99\linewidth]{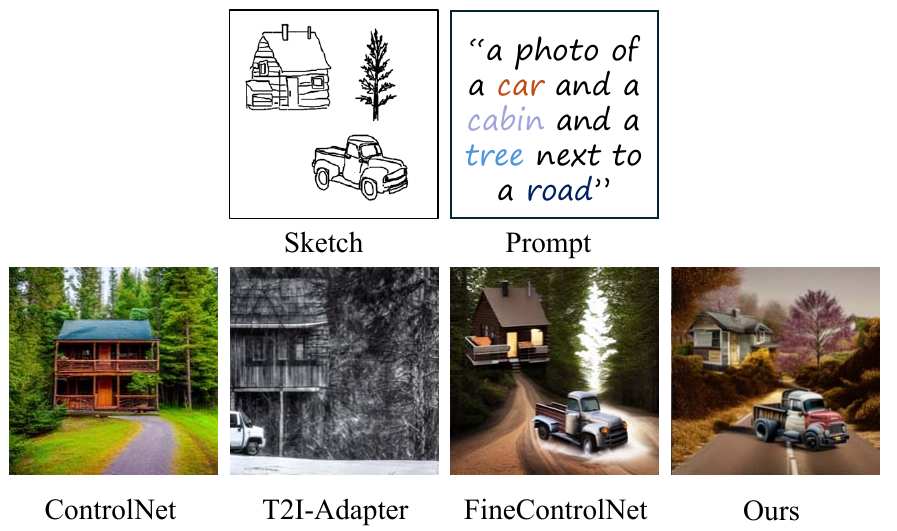}
    \caption{The existing sketch-guided text-to-image diffusion models perform poorly in generating scene sketches. ControlNet~\cite{zhang2023adding} and T2I-Adapter~\cite{mou2024t2i} generate images with object loss and background neglect. FineControlNet~\cite{choi2023finecontrolnet} can generate shapes from the sketch, but they may not always match the semantics, such as the ``tree'' in the image. Additionally, FineControlNet exhibits obvious segmentation between foreground and background, making it difficult to naturally blend them together.}
    \label{fig:issue}
\end{figure}

For the tasks of sketch-guided image generation, Sketch2Photo \cite{chen2009sketch2photo} treats it as a blending of multiple searched object images. However, the limitations of the filtering algorithm result in the model failing to filter the background in scenes and leading to synthesis failure. In addition, SketchyCOCO~\cite{gao2020sketchycoco} separates the foreground and background in sketches, using the generated foreground as a guide to generate the background. Nevertheless, SketchyCOCO's generalization is limited in the dataset as it only implements this method on 9 foreground classes. The state-of-the-art (SOTA) diffusion model-based approaches can usually perform well in generating individual objects, maintaining shape consistency and richness in details~\cite{zhang2023adding, mou2024t2i}. However, when describing the semantics of an entire scene using sketches, the content of the images may have diffculty in maintaining the integrity of objects and backgrounds (as shown in Figure~\ref{fig:issue}). FineControlNet~\cite{choi2023finecontrolnet} failed to naturally fuse foreground and background in generation. We observed that the coupling between foreground and background generation leads to semantic confusion, such as the sketch describing a tree being generated as blank space in the background forest. Due to the varying drawing abilities of users, hand-drawn sketches exhibit different levels of abstraction, lacking accurate 3D information and a coherent understanding of the scene. On the other hand, Diffusion models tend to generate realistic images. However, the conflict between preserving the sketch outlines and adhering to real-world semantics results in distorted and deformed generated images.

In this work, we propose the sketch-guided scene image framework based on the text-to-image diffusion model. As shown in Figure~\ref{fig:teaser}, to construct images across domains while maintaining the spatial distribution of the original scene sketch, we decouple the generation of foreground and background, breaking the image generation task into two subtasks: object-level cross-domain generation and scene-level image construction. In the object-level cross-domain generation, we utilize ControlNet to generate corresponding images from independent sketch objects, effectively preserving the shape of sketches and enriching the details. To address style conflicts in objects images and maintain consistent visual features in scene generation, we customize the detailed visual features of the generated objects with special identity tokens. Specifically, we use masks for the corresponding areas to ensure the model focuses only on pixels related to the objects while ignoring background influences. In the scene-level image construction, we guide the latent representation with foregrounds by combining the foreground images and masks. We first generate the background using a background prompt, and directly blend the foreground and background in the latent space. After determining the spatial distribution, we infer the entire image with global prompts containing special identity tokens. This process bridges the separation caused by previous blending and naturally fuses the foreground and background.

The main contributions of this work are listed as follows:
\begin{itemize}
    \item We propose a novel scene sketch-to-image generation method based on text-to-image diffusion model, achieving cross-domain generation of scene-level sketches while maintaining consistent spatial distributions with the sketch inputs. 
    \item We decouple the generation of the foreground and background, achieving a balance between foreground fidelity and foreground-background fusion during the final merging process.
    \item We validate both quantitatively and qualitatively that the proposed framework can accomplish the cross-domain scene sketch generation task. Compared to SOTA methods, our approach demonstrates higher generation quality and better sketch-image consistency.
\end{itemize}

\begin{figure*}
    \centering
    \includegraphics[width=0.99\linewidth]{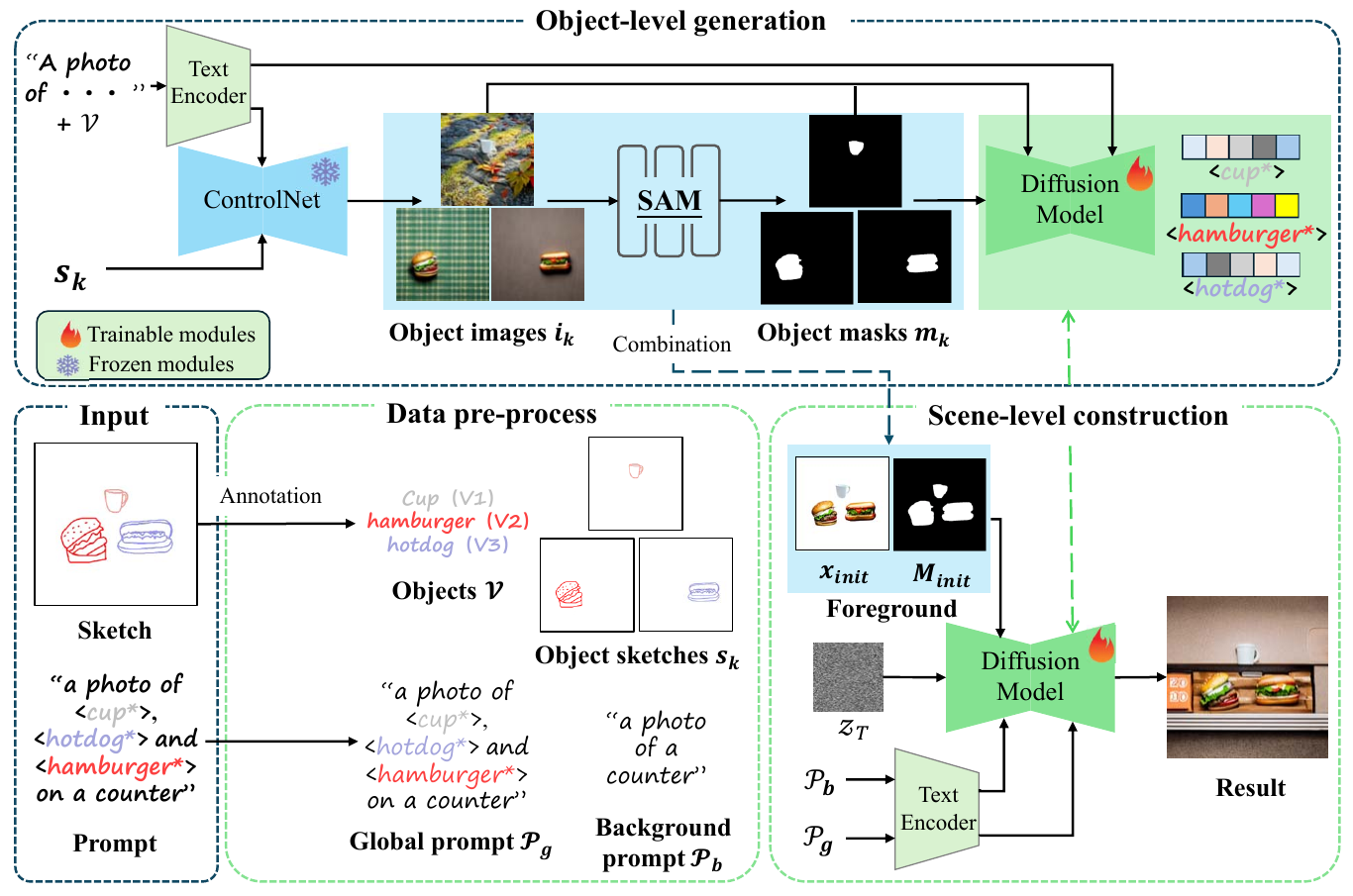}
    \caption{The proposed sketch-guided scene image generation framework consists of two main components: object-level generation and scene-level construction. (1) Object-level generation: given the input scene sketches, we annotate and separate individual object sketches and complete cross-domain object generation using ControlNet~\cite{zhang2023adding}. After generation, the images are segmented into masks, and Masked Diffusion Loss~\cite{avrahami2023break} is employed during training to reverse the visual features into unique identity embeddings. (2) Scene-level construction: In the trained diffusion model, we construct masks and initial foreground image that conform to the sketch space layout and guide the generation of the foreground during denosing process. We incorporate guidance in fewer inference steps, allowing the model greater freedom to iteratively refine and resolve scene inconsistencies between foreground and background.}
    \label{fig:framework}
\end{figure*}

\section{Related Work}

\subsection{Sketch-Based Image Generation}

Sketch presents an intuitive and flexible means of expression, but may exhibit different degrees of abstraction due to drawing skills. It is an important but challenging topic to generate various image contents from freehand sketches in computer graphics, such as the generation of human face\cite{deepfacedraw,peng24}, cartoon images~\cite{anifacedraw}, and dynamical effects~\cite{bo11,xie2024dualsmoke}. SketchyGAN~\cite{chen2018sketchygan} introduced an automatic sketch data augmentation and leveraged generative adversarial network (GAN) structure to accomplish sketch-to-image generation for 50 categories. SketchyCOCO~\cite{gao2020sketchycoco} proposed a foreground-background decoupling method for scene sketches and adopted EdgeGAN to separately generate the foreground and background images. 
Diffusion models can yield superior image quality compared to GANs. In text-to-image(T2I) diffusion models, researchers attempt to use sketches as additional conditional information to guide diffusion model generation. ControlNet~\cite{zhang2023adding} duplicated and trained an additional downsampling UNet network to input additional control guidance to the original weight-frozen network with diverse conditional generations including sketches. Similarly, T2I Adapter~\cite{mou2024t2i} introduced a lightweight structure to finely control generation using additional external control signals for the original T2I diffusion model. Additionally, Voynov et al.~\cite{voynov2023sketch} trained a pixel-wise multilayer perceptron to enforce consistency between intermediate images and input sketches, achieving object-level sketch cross-domain generation. However, these existing approaches often struggle with handling complex scene information and perform poorly on scene-level sketches.

\subsection{Scene Image Generation}

Scene image generation refers to creating or synthesizing images that depict complex scenes, which typically contain multiple objects, backgrounds, and interactions between elements. StackGAN~\cite{zhang2017stackgan} generates images from text using a coarse-to-fine approach: the first stage generates a primitive image with outlines, and the second stage refines and adds details. However, text descriptions often struggle to accurately and reasonably depict complex scenes, leading to the proposal of many multi-modal methods for scene image generation. Layout2Im\cite{zhao2019image} generates scene images from a rough spatial layout (bounding boxes and object categories) based on GANs. Sg2Im~\cite{zhao2019image} introduced scene graphs to infer objects and their relationships. Scene graphs are processed by Graph Convolution Network to predict bounding boxes and masks, and finally, a cascaded refinement network is used to convert the layout into an image. 
Although text-to-image diffusion models can produce high-quality results, they perform poorly on scene images. Due to the use of text embeddings, diffusion models face issues such as catastrophic object loss, attribute mixing, and spatial confusion. Attend-and-Excite~\cite{chefer2023attend} attempts to fully generate each object in the image by continually activating the cross-attention layers of each key object. GLIGEN~\cite{li2023gligen} adds layout guidance to the diffusion models to align the attention layers of each object with the input layout, thereby generating scene images with reasonable layouts. Attention-Refocusing~\cite{phung2023grounded} builds on GLIGEN by introducing self-attention loss and cross-attention loss, which maximally constrain the attention distribution of objects within the given bounding boxes. In this work, we aim to achieve better generation results from scene sketch inputs than the previous sketch-guided diffusion models.

\subsection{Controllable Text-to-Image Generation}

Text-to-image diffusion models~\cite{ho2020denoising,song2020denoising} have yielded remarkable results. However, text-only diffusion models encounter challenges like object omission and attribute confusion. Prompt-to-prompt~\cite{hertz2022prompt} explores the influence of cross-attention mechanisms on generated outputs, showcasing direct control over attributes and objects. 
Furthermore, GLIGEN~\cite{li2023gligen}, Multidiffusion~\cite{bar2023multidiffusion}, and Layout-guidance~\cite{chen2024training} all incorporate additional conditional inputs such as layout, mask, and keypoints, augmenting textual information by attention layers to convey precise control signals during the denoising process. Unlike algorithms that focus on attention mechanisms, ControlNet~\cite{zhang2023adding} , SGDM~\cite{voynov2023sketch}, InstructPix2Pix~\cite{brooks2023instructpix2pix}  gradually aligning latent features spatially with out-of-domain guided conditionings by adding additional network structures like MLP and U-Net, resulting in excellent results consistent withcondition inputs.
In addition, previous work~\cite{gal2022image,ruiz2023dreambooth,han2023svdiff, wei2023elite} has also addressed the issue of object changes in diffusion models. They used inversion methods to analyze the embedding space and treated specific visual concepts from reference images as unique identity markers, ensuring consistency of objects in subsequent diffusion model generations. Despite significant progress in single-concept customization, multi-concept customization remains a challenge. Custom Diffusion~\cite{kumari2023multi} attempts to combine new concepts, jointly training multiple concepts and merging fine-tuned models into one model through closed-loop constraint optimization. Additionally, Mix-of-show~\cite{gu2024mix} challenges multi-object concept customization using multiple Low-Rank Adaptation (LoRA)~\cite{hu2021lora}, employing embedding-decomposed LoRA to preserve intra-domain features of individual concepts and using region-controllable sampling to address issues such as attribute binding and object omission.  Break-A-Scene~\cite{avrahami2023break} extends from one scene image to multiple scene images, allowing random selection of different object combinations during scene training. 


\section{Method}

Our method aims to generate realistic images from scene sketches and is divided into two main components: object-level cross-domain generation and scene-level image construction as shown in Figure~\ref{fig:framework}. In the object-level cross-domain generation, we aim to accomplish cross-domain generation of all foreground objects and learn the corresponding image features. We first generate each object image from individual object sketches by ControlNet. After segmenting the object masks using SAM, we train the diffusion model to invert the visual features of the generated objects into identity embeddings (see Section~\ref{sec:4.1} and top of Figure~\ref{fig:framework}). In the scene-level image construction, we focus on generating the background and fusing the foreground with the background. We decouple the generation of the foreground and background, independently generate background while embedding the separately generated layout-compliant foreground objects into the latent space, then perform final inference to fuse the foreground and background. First, we combine all the foreground images and masks from the previous step as the initial guide. To decouple the generation of the foreground and background, background prompts without objects are used to infer the latent distribution of the background. Subsequently, we add noise to the initial foreground guidance and blend it into the corresponding positions. Once the layout is stabilized, we remove the foreground-background blending and introduce global prompts with unique identity tokens, allowing the diffusion model to freely infer a coherent image. This customized inference helps bridge inconsistencies between the foreground and background and fine-tune unrealistic parts of the sketch, resulting in a suitable image (see Section~\ref{sec:4.2} and bottom of Figure~\ref{fig:framework}).

\subsection{Preliminary: Diffusion Models}
\label{sec:3.1}

Latent Diffusion Model (LDM)~\cite{rombach2022high} trained an AutoEncoder, including an encoder $\mathcal{E}$ and a decoder $\mathcal{D}$. After the image $x$ is compressed by the encoder $\mathcal{E}$ to latent representation $z$, the diffusion process is performed on the latent representation space. Given a latent sample $z_0$, the Gaussian noise is progressively increased to the data sample during $T$ steps in the forward process, producing the noisy samples $z_t$, where the timestep $t=\{1, \ldots, T\}$. As $t$ increases, the distinguishable features of $x_0$ gradually diminish. Eventually when $T \rightarrow \infty$, $x_T$ is equivalent to a Gaussian distribution with isotropic covariance. Finally, LDM infers the data sample $z$ from the noise $z_T$ and $\mathcal{D}$ restores the data $z$ to the original pixel space and gets the result images $\widetilde{x}$. In the training process, the loss is defined as:
\begin{equation}
    \label{equ:ldm}
    L_{L D M}:=\mathbb{E}_{\mathcal{E}(x), \epsilon \sim \mathcal{N}(0,1), t}\left[\left\|\epsilon-\epsilon_\theta\left(z_t, t\right)\right\|_2^2\right]
\end{equation}
where $\epsilon$ is the sample noise from normal distribution and $\epsilon_\theta$ is   the model's predicted noise. $\mathbb{E}_{\mathcal{E}(x), \epsilon \sim \mathcal{N}(0,1), t}$ means the evidence lower bound (ELBO).

Blended Latent Diffusion (BLD)~\cite{avrahami2023blended} is based on text-to-image LDM, and proposes a method for text-driven image editing that retains the original image pixels and infers the latent representation of the masked areas. For a given image $x$ and mask $M$, BLD encodes the image into the latent space as $z_{init} \sim \mathcal{E}(x)$ and downsamples the mask as $m_{latent}$ to the same spatial dimensions for blending. During each iterative denoising step of the diffusion process, BLD obtains the latent representation $z_{bg}$ of the background by noise-corrupting the original latent $z_{init}$ to the corresponding noise level, while using the guiding text prompt $d$ as a condition to obtain a less noisy latent foreground $z_{fg}$. Finally, the mask is used to blend the two latent representations:
\begin{equation}
    \label{bld}
    z_t \leftarrow z_{bg} \odot (1 - m_{latent}) + z_{fg} \odot m_{latent}
\end{equation}
where $\odot$ is element-wise multiplication. The image outside the mask is enforced to remain unchanged, while the pixels within the mask adhere to the text-based inference. 

The separation of foreground and background mentioned in BLD has sparked our interest. In the scene sketches proposed by SketchyCOCO~\cite{gao2020sketchycoco}, the foreground and background can typically be separated, with the foreground often influencing the generation of the background. Additionally, people tend to focus more on the details of the foreground, while leaving the background blank or roughly sketched in scene sketches. In Section~\ref{sec:4.2}, we aim to utilize blending method to decouple the generation of the foreground and background, inferring a reasonable background while preserving the details of the foreground and maintaining stylistic consistency between the foreground and background.

\begin{figure}
    \centering
    \includegraphics[width=0.99\linewidth]{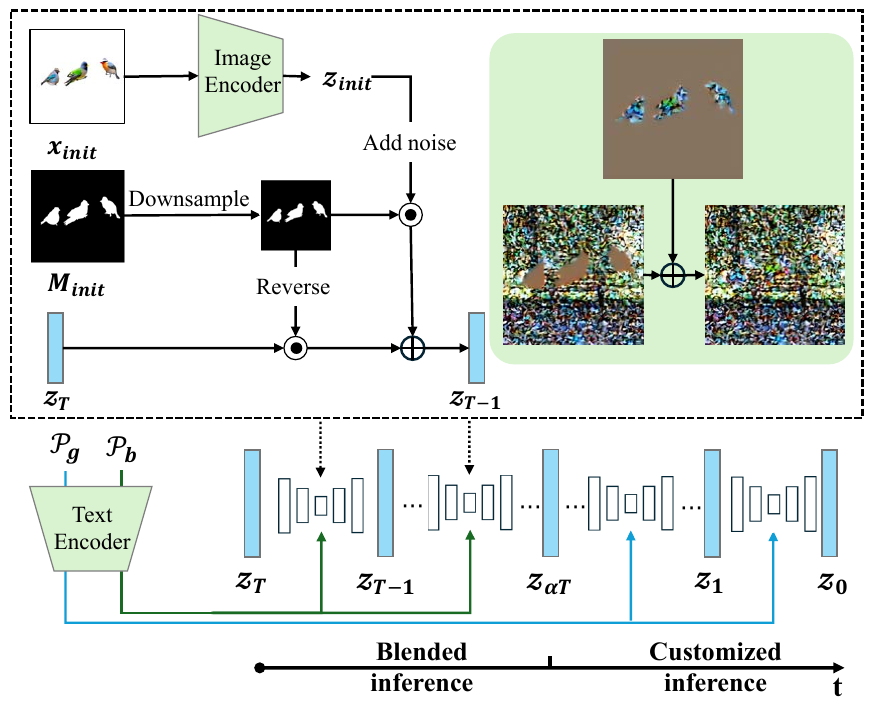}
    \caption{The inference process in our method. In the blended inference process, background prompt $\mathcal{P}_b$ will be utilized to inference the background. The foreground image $x_{init}$ is encoded and noised to represent the foreground objects, mask $M_{init}$ is used to blend the latent representations of the foreground and background. In customized inference, we use a global prompt $\mathcal{P}_g$ containing special identity tokens to guide the model in generating images from the blended latent representations.}
    \label{fig:inference}
\end{figure}

\subsection{Object-Level Generation}
\label{sec:4.1}

In the object-level generation process, we aim to generate detailed images from individual sparse object sketches, avoiding semantic confusion and identity loss that may arise in scene-level generation. As shown in the top of Figure~\ref{fig:framework}, we annotate and separate individual object sketches $\{$$s_1$, \ldots ,$s_k$$\}$ with individual prompts $\{$$p_1$, \ldots ,$p_k$$\}$ from the given scene sketches $\mathcal{S}$, where $k$ is the number of objects in the scene sketch. Sketches and prompts are processed through a pre-trained ControlNet~\cite{zhang2023adding} to generate a series of images $\{$$i_1$, \ldots ,$i_k$$\}$, each containing a single object. 

To preserve the model's inference capabilities rather than merely pasting pixels, we invert the corresponding objects into unique identity embeddings $\{$$v_1$, \ldots ,$v_k$$\}$ to retain the generated visual features. We use the masked diffusion loss~\cite{avrahami2023break} to precisely understand the concepts or objects without the background influence. Masked diffusion loss directs the model's attention to the desired masks, thus resolving ambiguity in training objectives. Specifically, we extract the corresponding masks $\{$$m_1$, \ldots ,$m_k$$\}$ from object images by Grounded Segment Anything Model~\cite{ren2024grounded}. During the training process of the diffusion model, the latent representation $z_t$ corresponding to each timestep $t$ is penalized only for pixels covered by the respective mask $m_i$. The loss is defined as follows:

\begin{equation}
    \mathcal{L}_{rec}=\mathbb{E}_{z, s, \epsilon \sim \mathcal{N}(0,1), t}\left[\left\|\epsilon \odot m_i-\epsilon_\theta\left(z_t, t, p_i\right)\odot m_i\right\|_2^2\right]
    \label{equ:mdl}
\end{equation}
where $p_i$ is the text prompt, $\epsilon$ is the added noise and $\epsilon_\theta$ is the denoising network. By using the masked diffusion model, the model is compelled to focus exclusively on the regions covered by the mask, faithfully reconstructing the target concepts, thus eliminating the influence of other pixels on customized learning.

\begin{algorithm}[t]
	\caption{Scene-level construction: given\\ a Stable Diffusion Model \{\textit{VAE} = $(\mathcal{E}(x),\mathcal{D}(z))$, \\ \textit{DiffusionModel} = (\textit{noise}$(z,t)$, \textit{denoise}$(z,\mathcal{P},t))\}$, which trained in Section~\ref{sec:4.1}.} 
	\label{alg3} 
	\begin{algorithmic}[1]
		\REQUIRE initial foreground image $x_{init}$, initial foreground mask $M_{init}$, diffusion steps $T$, global prompt $\mathcal{P}_g$ \\and background prompt $\mathcal{P}_b$.
		\ENSURE generated image $\hat{x}$ that conforms to the objects and layout of the scene sketch $\mathcal{S}$.
		\STATE $z_{init} \sim \mathcal{E}(x_{init})$
            \STATE $m_{init} = downsample (M_{init})$
            \FOR{$t$ from $T$ to 0}
                \IF{$t > \alpha T$}
                \STATE $z_{bg} \sim denoise(z_t, \mathcal{P}_b, t)$
                \STATE $z_{fg} \sim noise(z_{init}, t)$
                \STATE $z_t \leftarrow z_{bg} \odot (1 - m_{init}) + z_{fg} \odot m_{init}$
                \ELSE
                \STATE $z_{t} \sim denoise(z_t, \mathcal{P}_g, t)$
                \ENDIF
            \ENDFOR
            \STATE $\hat{x} = \mathcal{D}(z_0)$
            \RETURN $\hat{x}$
	\end{algorithmic} 
\end{algorithm}

\begin{figure*}[htbp]
    \centering
    \includegraphics[width=0.99\linewidth]{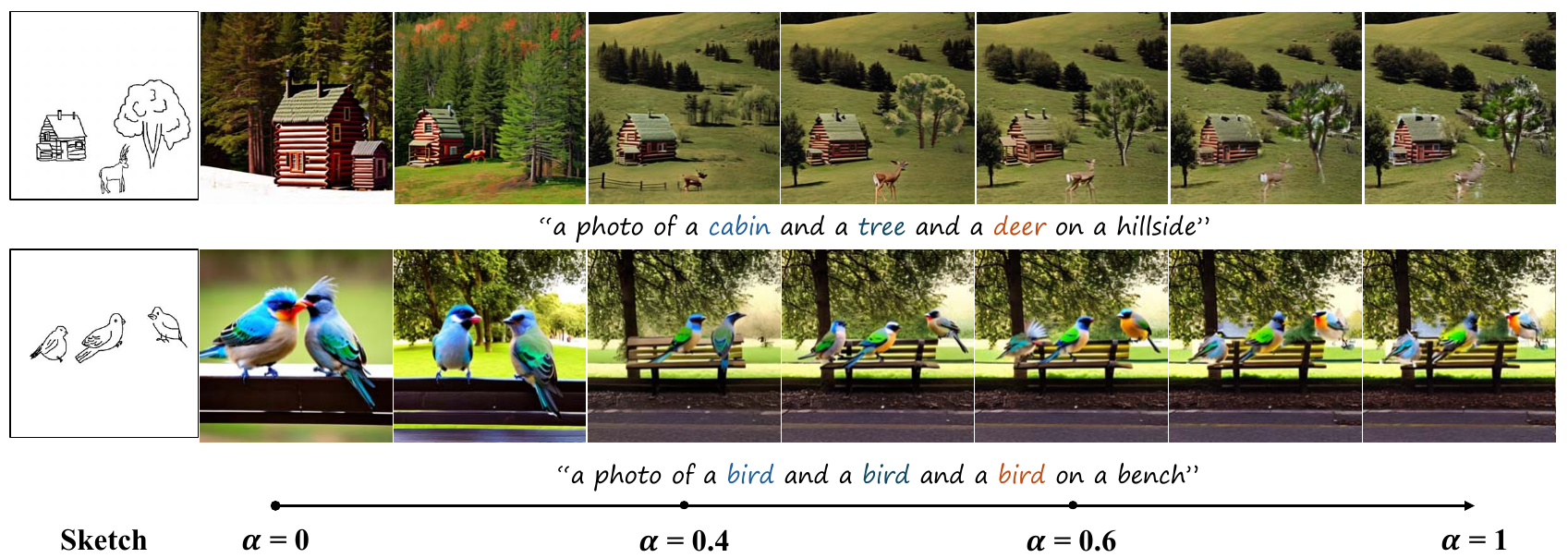}
    \caption{The generated results with different $\alpha$. $\alpha = 0$ means without the blended inference and $\alpha = 1$ represent the full blending during the inference process. We observed that the balance between layout accuracy and foreground-background consistency can be achieved within the range of 0.4 to 0.6.}
    \label{fig:t}
\end{figure*}

\subsection{Scene-Level Construction}
\label{sec:4.2}

After obtaining the foreground images, we proceed with background image generation and scene-level construction. Our goal is to embed foreground objects into their corresponding spatial distributions without interfering with background generation, and blend foreground and background while maintaining consistency and smooth transition. Our approach is summarized in Algorithm~\ref{alg3}, and depicted in Figure~\ref{fig:inference}.

Inference is performed using the diffusion model trained in Section~\ref{sec:4.1}. Due to directly blending latent representations at all time steps will lead to a noticeable segmentation between foreground and background. We utilize hyperparameters $\alpha \in [0,1]$ to divide the inference process into two parts, blended inference and customized inference, and prepare corresponding prompts for each. The global prompt $\mathcal{P}_g$ includes previously learned identity tokens and background, that describe the objects in the scene sketch and the typically overlooked background. The background prompt $\mathcal{P}_b$ focuses only on describing the background, omitting all foreground objects. For example, the $\mathcal{P}_g$ is ``a photo of a chair and table in a room'', the $\mathcal{P}_b$ will be ``a photo of a room''.

\textit{\textbf{Blended inference.}} Since the positions of the objects have been determined early in the diffusion process. When $t > \alpha T$, we provide the $x_{init}$ to the diffusion model for layout guidance. We implement the blending process aim to embed the initial foreground image into the latent representation without affecting the generation of the background. Specifically, we construct the pixels of each objects into an initial foreground image $x_{init}$ and mask $M_{init}$ from the generated objects images and masks. Note, the $M_{init}$ is differs from $M$ in BLD, $M_{init}$ represents the mask area where the foreground pixels need to be retained, while $(1 - M_{init})$ indicates the background area that the model needs to inference. To provided the initial objects layout guidance, $x_{init}$ is encoded to $z_{init}$ and added noise to form $z_{fg}$. $M_{init}$ is downsampled to $m_{init}$ to represent the region where the foreground is located. Similarly, derive the initial latent representation $z_t$ containing the background from the background prompt $\mathcal{P}_b$ to generate the background individually. The latent representation $z_{bg}$ is predicted and denoised form $z_t$ and $\mathcal{P}_b$ in the timestep $t$. We employ $z_{fg}$ and $z_{bg}$ to compose the latent representation $z_t$: 
\begin{equation}
    \label{equ:zt}
        z_t \leftarrow z_{bg} \odot (1 - m_{init}) + z_{fg} \odot m_{init},\quad t > \alpha T
\end{equation}
Where the foreground shape and position align with the guidance of $x_{init}$, while the background is inferred from the background prompt $\mathcal{P}_b$.

\textit{\textbf{Customized inference.}} When $t \leq \alpha T$, we leverage the model's inference capability to denoise without the foreground image guidance. Although the background and foreground can each appear intact in the latent representation during the blending process, there is still a noticeable segmentation between them. Global prompts $\mathcal{P}_g$ are introduced to guide the model with the entire semantic of the scene, encouraging it to denoise step by step from the noise and bridge the gap between foreground and background, generating images with natural transitions. Additionally, because the global prompt contains the trained identity tokens, the model infers objects that align with the pre-trained visual features rather than generating new objects from scratch.  The final generated objects will have contours and details consistent with the sketch. At this point, we update the latent representation to:
\begin{equation}
\label{free}
    z_{t} \sim denoise(z_t, \mathcal{P}_g, t),\quad t \leq \alpha T
\end{equation}
Under the model's autonomous inference, significant reconciliation of foreground-background conflicts is achieved, and minor adjustments are made to bring objects closer to reality without altering their details.

\section{Experiments and results}

We utilize the pre-trained ControlNet Scribble model~\cite{zhang2023adding} to generate object images and Stable Diffusion V-2.1~\cite{rombach2022high} to train the customized identity tokens. The Grounded SAM~\cite{ren2024grounded} is utilized as our segmentation model. We constructed dataset, contained 30 scene sketches with 57 classes, composed from the Sketchy dataset~\cite{sangkloy2016sketchy}, each scene sketch containing 2 to 4 objects. We prepared 35 common backgrounds, such as `` on the road'', ``on the hillside'', and ``in the square''. The generation seed was randomly sampled from the range of $[0,50]$, inference timestep is $50$ and $\alpha$ sampled from $[0.4,1]$. 

We select the pre-trained ControlNet Scribble~\cite{zhang2023adding},  FineControlNet~\cite{choi2023finecontrolnet} and T2I-Adapter~\cite{mou2024t2i} as the benchmarks. Among them, the inputs for the ControlNet and FineControlNet are the scene sketches and the prompts, we replace the identity tokens in our method with the corresponding class labels. In particular, the inputs for the T2I-Adapter consists of our annotated object sketches and class labels. All input images are $512\times512$ in resolution.

\subsection{Qualitative Experiments}

\begin{figure*}[htbp]
    \centering
    \includegraphics[width=0.99\linewidth]{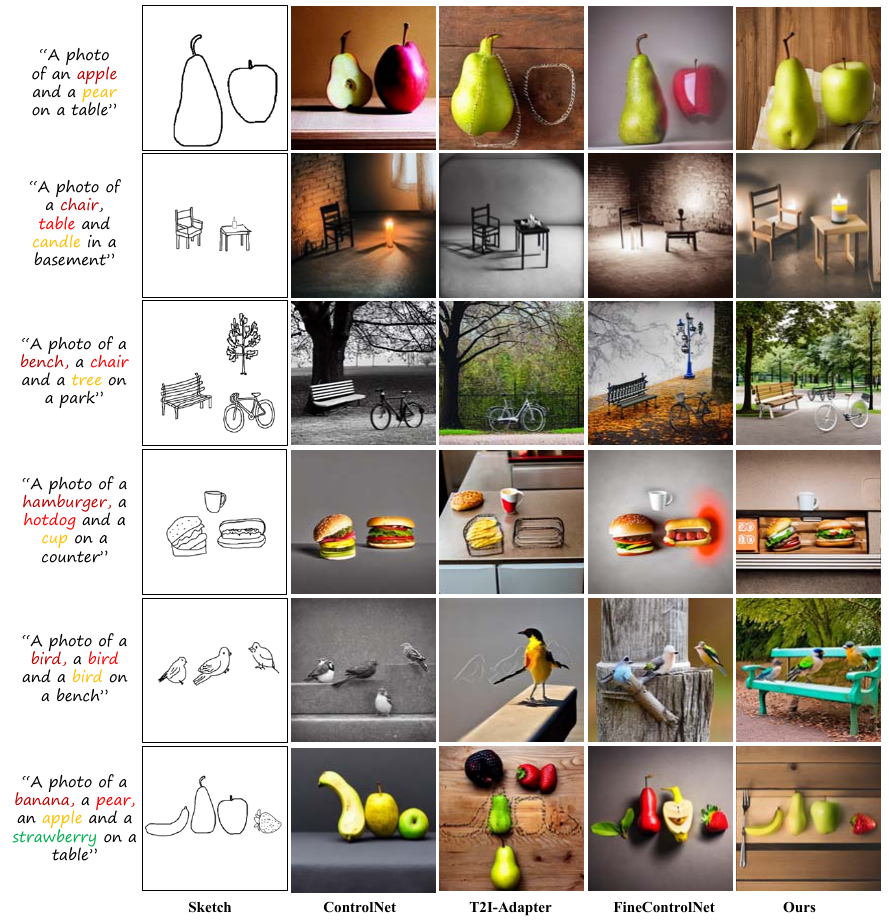}
    \caption{A qualitative comparison between our method and ControlNet~\cite{zhang2023adding}, T2I-Adapter~\cite{mou2024t2i} and FineControlNet~\cite{choi2023finecontrolnet}. As can be seen, ControlNet and T2I-Adapter struggle with preserving the objects in scene sketches. FineControlNet preserves the objects better than ControlNet and T2I-Adapter, but always ignores background generation. Finally, our method is able to generate objects following the sketch guidance and generate the background based on prompts.}
    \label{fig:com}
\end{figure*}

We first verified the impact of different $\alpha$ values on the generated results. As shown in the Figure~\ref{fig:t}, $\alpha = 1$ represents that removing the blended inference, while $\alpha = 0$ represents the implementation of blended inference throughout the entire inference process. When $\alpha$ is small, it is often difficult to maintain the layout guidance from the sketch. Conversely, there are often noticeable gaps between the foreground and background when $\alpha$ is large. A balance between maintaining layout stability and blending the foreground and background is often achieved when $\alpha \in [0.4,0.6]$.

We also report a qualitative comparison between our method and ControlNet, FineControlNet and T2I-Adapter. As shown in Figure~\ref{fig:com}, both ControlNet and T2I-Adapter struggle to fully generate the objects in the input scene sketches, with severe issues of object loss and semantic confusion. FineControlNet ensures that objects are generated in their respective positions as much as possible. However, the complex semantics often cause the model to ignore background generation, failing to adhere to the prompt semantics. Our method decouples foreground and background generation, faithfully generating foreground images from the sketches while also focusing on background generation, ultimately producing images that conform to both the input sketch and the prompt.

\subsection{User Preference Experiments}

We conducted a user preference experiment, where we provided each of the 102 participants with five sets of generated result images along with their corresponding sketches. We solicited three key opinions from the users in the form of a questionnaire: object consistency between images and sketches, background consistency between images and text, and the overall preference level of users for the generated images in each set. The object consistency and background consistency were rated on a five-point scale, with 1 indicating the lowest and 5 indicating the highest level of consistency. The image quality was rated on a single-choice, with the best quality image selected. As shown in the table~\ref{table:com}, our method demonstrates excellent capability in generating both objects and backgrounds, while being more preferred by users in overall image fusion.


\begin{table}[t]
\caption{User preference experiments between the ControlNet, T2I-Adapter, FineControlNet and our method. 
In the table, ``O-C'' represents object consistency, ``B-C'' represents background consistency, and ``T2I'' represents T2I-Adapter.}
\begin{center}
\begin{tabular}{cccc}
\toprule 
&O-C & B-C & Preference \\ \hline
ControlNet & 3.312 & 3.360 & 24.508\%\\ 
T2I& 2.534& 2.796 & 10.978\% \\
FineControlNet& 3.014 & 2.804  &10.782\%\\
Ours& \textbf{3.758}&\textbf{3.636}&\textbf{53.726\%}\\
\bottomrule 
\end{tabular}
\label{table:com}
\end{center}
\end{table}



\subsection{Ablation Study}

\begin{figure*}[htbp]
    \centering
    \includegraphics[width=0.99\linewidth]{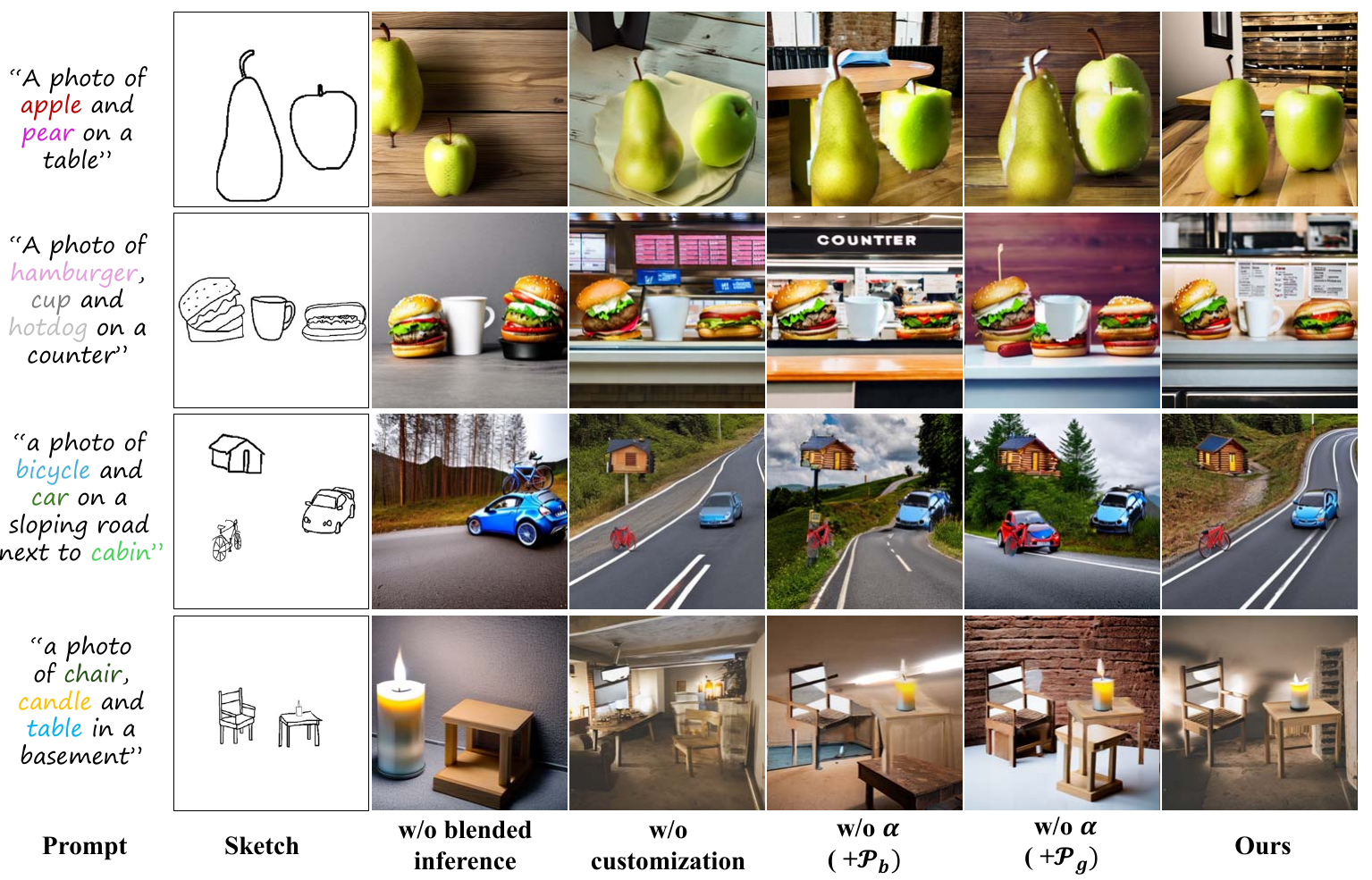}
    \caption{Qualitative ablation: we ablate our method by (a) removing blended inference, (b) removing customized training, (c) removing $\alpha$ with background prompt and (d) removing $t$ with global prompt. As can be seen, the model lose the layout guidance from sketch when removing blended inference and weakened the outline guidance when removing customized training. Without the $\alpha$, the model tends to thoroughly segment foreground and background, simply involve pixel-wise overlay.}
    \label{fig:ablation}
\end{figure*}

We conducted an ablation study, which includes removing the identity embeddings that trained in Section~\ref{sec:4.1}, removing the blended inference that introduced in Section~\ref{sec:4.2}, and removing the $\alpha$ (that means use the blended inference during the whole inference process). Since we have global prompt and background prompt, we removing $\alpha$ with these two prompts individually. Specifically, in the generation process, we use trained special identity tokens to refer to objects. When removing identity embeddings, we replace the identity tokens with the corresponding class labels.

\begin{figure*}
    \centering
    \includegraphics[width=0.99\linewidth]{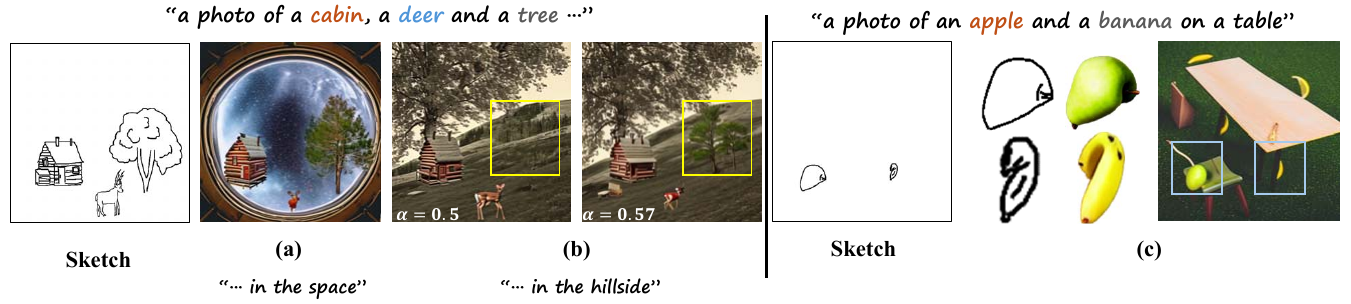}
    \caption{Limitations in our method. (a) When the composed scene does not align with a real-world scene, the generated images become semantically confused and fail to convey the correct content. (b) Although providing additional layout guidance can mitigate the issue of object disappearance, it cannot completely prevent this problem from occurring. (c) Overly abstract sketch objects are difficult to generate correctly in the image.}
    \label{fig:limitation}
\end{figure*}

As shown in the second column in Figure~\ref{fig:ablation}, when removing the blended inference, the generated results fail to adhere to the layout of the input sketch. Additionally, without the initial pixel guidance, issues such as identity blending and object loss become prominent. In the third column, when removing customized training, although the guided layout is still maintained, the control over object details is lost. Due to the blending with reference foreground images, objects generally conform to the input sketch in terms of scene layout. However, the contours and details of individual objects often do not match the sketch. such as the ``apple'' in the first row and the ``cup'' in the second row. Although the foreground images in the blended inference provide the model with initial guidance, in the subsequent customized inference stage, non-customized models often fail to maintain the style of foreground objects completely unchanged and tend to modify details during denoising. We also attempted to perform the blending process separately using global prompts and background prompts throughout the entire inference process, as shown in the fourth and fifth columns. Regardless of which prompts were used, there is a clear disconnect between the foreground and background. When using background prompts, complete foreground and background are generated but they do not fused together naturally, such as the houses flying in the sky in the third row. When using global prompts, additional objects are generated unnecessarily, leading to issues like object loss and identity blending. For example, in the second row, almost all objects have features resembling hamburgers.

\section{Conclusion}

In this paper, we explore the challenges of cross-domain scene image generation from scene sketches using text-to-image diffusion models. We aim to generate objects that adhere to the layout of the scene sketch while ensuring the background aligns with the semantic description. Our approach can achieve a balance between object consistency and foreground-background fusion by dividing the inference process into two parts. Ablation experiments demonstrate the effectiveness of each component of our proposed method. In addition, we conduct both qualitative and quantitative experiments to show that our approach outperforms SOTA sketch-guided diffusion models.

Our proposed method may suffer some limitations. When the foreground and background fail to semantically align with real-world scenes, the generated images may exhibit distortions (as shown in Figure~\ref{fig:limitation} (a)). Although our method provides spatial guidance through layout during the generation process, the issue of object loss in diffusion models persists (see the yellow boxes in Figure~\ref{fig:limitation} (b)). Specifically, this issue becomes more severe as the number of objects in the image increases, choosing appropriate values for $\alpha$ and the generation seed can improve this situation. 
The bad-drawn rough sketches may make it challenging to construct reasonable individual objects, and difficult to fuse them seamlessly into the background during generation.
As shown in Figure~\ref{fig:limitation} (c), the abstract representations of apples and bananas ultimately resulted in image distortion, producing two redundant color patches under the table in the background (highlighted in the blue boxes).

{\small
\bibliographystyle{ieee_fullname}
\bibliography{sketch2scene}
}

\end{document}